\documentclass[conference]{IEEEtran}

\usepackage{cite}
\usepackage{amsmath}
\usepackage{amsmath,amssymb,amsfonts}
\usepackage{amssymb}
\usepackage{graphicx}
\usepackage{wrapfig}
\usepackage{multirow}

\begin{document}

\bstctlcite{IEEEexample:BSTcontrol} % Control bibliography options

\title{Fusion Intelligence: Confluence of Natural and Artificial Intelligence for Enhanced Problem-Solving Efficiency}

% use a multiple-column layout for up to three different
\author{\IEEEauthorblockN{Rohan Reddy Kalavakonda$^{1,*}$, Junjun Huan$^{1}$, Peyman Dehghanzadeh$^{1}$, Archit Jaiswal$^{1}$, \\
Soumyajit Mandal$^{2}$, and Swarup Bhunia$^{1}$}
\IEEEauthorblockA{$^{1}$Department of Electrical and Computer Engineering, University of Florida, Gainesville, FL 32611\\
Email: rohan.reddykalav@ufl.edu}
\IEEEauthorblockA{$^{2}$Instrumentation Department, Brookhaven National Laboratory, Upton, NY 11973}}


\maketitle{\thispagestyle{plain}}
\pagestyle{plain}

\begin{abstract}This paper introduces Fusion Intelligence (FI), a bio-inspired intelligent system, where the innate sensing, intelligence and unique actuation abilities of biological organisms such as bees and ants are integrated with the computational power of Artificial Intelligence (AI). This interdisciplinary field seeks to create systems that are not only smart but also adaptive and responsive in ways that mimic the nature. As FI evolves, it holds the promise of revolutionizing the way we approach complex problems, leveraging the best of both biological and digital worlds to create solutions that are more effective, sustainable, and harmonious with the environment. We demonstrate FI's potential to enhance agricultural IoT system performance through a simulated case study on improving insect pollination efficacy (entomophily).
\end{abstract}

\IEEEpeerreviewmaketitle

\section{Introduction}
\label{sec:introduction}

Artificial Intelligence (AI) has revolutionized the way machines interact with the physical world, mirroring some facets of human cognition. The integration of AI with sensory inputs and actuation mechanisms has given rise to autonomous systems capable of self-regulation and decision-making in real-time. These systems are designed to process and analyze data streams from sensors, which act as proxies for human sensory organs, allowing the AI to perceive its environment. The computational models underlying AI are built upon algorithms that learn and evolve, drawing from the principles of machine learning and neural networks. This enables AI to recognize complex patterns, make informed decisions, and carry out tasks with a degree of autonomy that was previously unattainable. However AI necessitates considerable computational resources, particularly for advanced algorithms like deep neural networks~\cite{wu2022sustainable}. The computational demand escalates when training models on extensive datasets. Moreover, the precision of AI outputs is linked to the quality of sensor data, which poses its own set of challenges. Sensor calibration and the assurance of data accuracy are critical, especially in environments where conditions fluctuate unpredictably, impacting the reliability of the data collected.

Natural systems like insects provide a benchmark for improving AI systems. Biological sensory systems excel in capturing and processing environmental stimuli. While biomimetic sensors mimic perceptions of light, sound, and odor\cite{jung2019bioinspired}, they struggle with accuracy, precision, range, and robustness. Biological systems use attention mechanisms to focus on important stimuli and respond quickly to environmental cues. Animals from simple insects to complex primates display a range of cognitive abilities involving instincts, conditioning, and learning. Replicating these abilities in AI systems involves not only the sensory processing but also the integration of this information with motor functions, which in natural systems is seamlessly efficient and adaptable. In contrast, artificial sensor systems and AI algorithms are energy-inefficient and need vast memory/data to replicate a fraction of these abilities, making it challenging to deploy biomimetic AI on edge devices.

\begin{figure*}[htbp]
\centering
\includegraphics[width=0.9\textwidth]{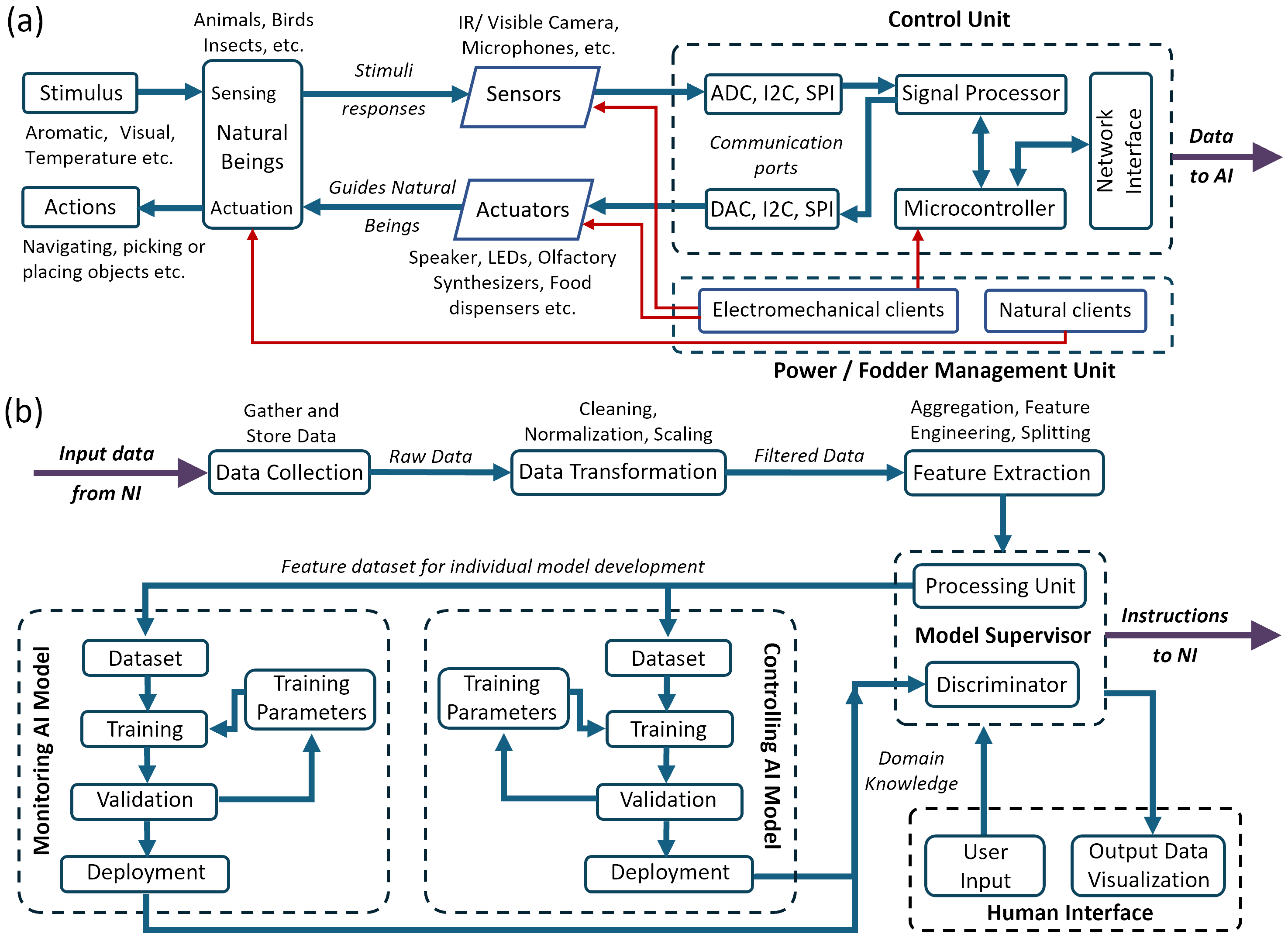}
\caption {Flow chart of the NI and AI subsystems used within the FI system architecture. (a) The NI subsystem hosts the biological entities that act as sensors and actuators in the environment. Compatible electromechanical sensors observe these entities. The control unit digitizes the sensor outputs, sends them to the AI subsystem, and decodes its instructions. The latter are relayed to the entities via compatible actuation devices like speakers or food dispensers. (b) The AI subsystem hosts the digital processes required to interpret the actions of biological entities and alter their behaviour based on user objectives. The Supervisor compares the observed and required outcomes to manage the AI model outputs in real-time based on user-defined control algorithms.}
\label{fig:fiarchflow}
\end{figure*}

In this paper we propose Fusion Intelligence(FI), a bio-inspired integration framework which merges the intrinsic capabilities of natural systems with the potential of AI-driven electronic systems. An overview of the proposed system is shown in Fig.~\ref{fig:fiarchflow}. FI can be applied in various domains including agriculture, environmental monitoring, search and rescue operations, national defence, and security. An agricultural application of the FI framework was evaluated in a simulated virtual environment and is presented as a case study. The study demonstrates an FI system with bees acting as sensors and actuators to enhance pollination across a large field. By integrating AI with NI, the system learns from the insects and then guides them to optimize pollination.

\section{Related Work}
\label{sec:background}

\begin{table*}[t]
\caption{Comparison with Earlier Work}
\label{tab:comp}
\resizebox{\textwidth}{!}{%
\begin{tabular}{|l|l|l|l|l|l|l|}
\hline
\multicolumn{1}{|c|}{\textbf{Previous Works}} & \multicolumn{1}{c|}{\textbf{Main Focus}}                                                      & \multicolumn{1}{c|}{\textbf{Insect Used}} & \multicolumn{1}{c|}{\textbf{Key Features}}                                                                     & \multicolumn{1}{c|}{\textbf{\begin{tabular}[c]{@{}c@{}}Complexity of \\ Integration\end{tabular}}}                  & \multicolumn{1}{c|}{\textbf{\begin{tabular}[c]{@{}c@{}}Effect on \\ Daily Routine\end{tabular}}}               & \multicolumn{1}{c|}{\textbf{\begin{tabular}[c]{@{}c@{}}Extent of \\ Damage/Invasiveness\end{tabular}}}      \\ \hline
Sato et al 2008.~\cite{Sato2008}              & \begin{tabular}[c]{@{}l@{}}Implantable   \\ flight control\end{tabular}                       & Cotinis texana                            & \begin{tabular}[c]{@{}l@{}}Neural/muscular  stimulators, \\ visual stimulator, \\ microcontroller\end{tabular} & \begin{tabular}[c]{@{}l@{}}Complex;   involves\\  multiple implantations\end{tabular}                               & \begin{tabular}[c]{@{}l@{}}Not discussed, but\\ likely affected due to\\ physical modifications\end{tabular} & \begin{tabular}[c]{@{}l@{}}Invasive; requires \\ precise implantation \\ during pupal stage\end{tabular}    \\ \hline
Sato et al 2009~\cite{Sato2009}               & \begin{tabular}[c]{@{}l@{}}Radio-frequency\\ neural control\end{tabular}                      & Mecynorhina spp.                          & \begin{tabular}[c]{@{}l@{}}RF   receiver, microbattery, \\ electrode stimulators\end{tabular}                  & \begin{tabular}[c]{@{}l@{}}Complex;   integration \\ of RF system and \\ electrodes\end{tabular}                    & \begin{tabular}[c]{@{}l@{}}Alters normal behavior\\  for controlled flight\end{tabular}                        & \begin{tabular}[c]{@{}l@{}}Invasive; involves \\ implantation of\\  electrodes and\\  receiver\end{tabular} \\ \hline
Kakeiet al.~\cite{Kakei2022}                  & \begin{tabular}[c]{@{}l@{}}Organic solar cell\\ on cyborg insects\end{tabular}                & G. portentosa                             & \begin{tabular}[c]{@{}l@{}}Ultrathin   organic solar cell, \\ wireless locomotion control\end{tabular}         & \begin{tabular}[c]{@{}l@{}}Moderately   complex;\\  attachment of solar cell\\  and electronics\end{tabular}        & \begin{tabular}[c]{@{}l@{}}Alters normal behavior\\  for controlled \\ movement\end{tabular}                   & \begin{tabular}[c]{@{}l@{}}Minimally invasive; \\ attachment of \\ components on \\ exterior\end{tabular}   \\ \hline
Latif et al.~\cite{Latif2012}                 & \begin{tabular}[c]{@{}l@{}}Wirelessly navigated \\ biobots\end{tabular}                       & G. portentosa                             & \begin{tabular}[c]{@{}l@{}}ZigBee-enabled  \\  neurostimulation backpack\end{tabular}                          & \begin{tabular}[c]{@{}l@{}}Moderately  complex; \\ integration of wireless \\ navigation system\end{tabular} & \begin{tabular}[c]{@{}l@{}}Alters normal behavior \\ for controlled \\ movement\end{tabular}                   & \begin{tabular}[c]{@{}l@{}}Minimally invasive; \\ backpack mounted \\ externally\end{tabular}               \\ \hline
Bozkurt et al.~\cite{Alper2007}               & \begin{tabular}[c]{@{}l@{}}Microsystem \\ platform for flight\\ muscle actuation\end{tabular} & Manduca sexta                             & \begin{tabular}[c]{@{}l@{}}Microprobe-based platform \\ for muscle actuation\end{tabular}                      & \begin{tabular}[c]{@{}l@{}}Complex; implantation \\ of microprobes in \\ muscles\end{tabular}                       & \begin{tabular}[c]{@{}l@{}}Alters normal behavior \\ for controlled flight\end{tabular}                        & \begin{tabular}[c]{@{}l@{}}Invasive; precise \\ insertion during \\ early metamorphosis\end{tabular}        \\ \hline
FI case study                                 & \begin{tabular}[c]{@{}l@{}}Improving \\ pollination and \\ health of bees\end{tabular}        & Honey bees                                & Utilizes AI to assist bees                                                                                     & Relatively simple                                                                                               & \begin{tabular}[c]{@{}l@{}}Normal behavior is \\ not altered and can\\  be enhanced\end{tabular}               & \begin{tabular}[c]{@{}l@{}}Non-invasive; does\\  not involve\\  implantation\end{tabular}                   \\ \hline
\end{tabular}%
}
\end{table*}

For decades, insects have been studied for applications in national defense, agriculture, navigation, search and rescue, and explosives detection~\cite{Sato2008, Alper2007, Abdel-Raziq2021, Alper2016, Filipi2021, Habib2007}. Previous approaches involve integrating miniaturized sensors on insect bodies, such as wireless cameras~\cite{Vikram2020}, RFID tags~\cite{Henry2012}, harmonic radar~\cite{Osborne1999}, Lidar~\cite{Rydhmer2022}, auditory sensors~\cite{Fishel2021}, and flight recorders~\cite{Abdel-Raziq2021}. One alternative to direct control of insects for performing tasks is sensory control through stimuli, guiding insect activities~\cite{Rains2008, Filipi2021, Piqueret2022}, with insects trained via classical conditioning~\cite{Rains2008}, although this requires periodic reconditioning.

Table~\ref{tab:comp} compares previous work on controlling insects to our case study on using FI to improve bee pollination. All earlier works require a payload ranging from 3~mm to 30~mm long and weighing  422~mg to 8140~mg. However, a honey bee only measures 5.5-23~mm in length and has a payload capacity ($\approx$50$\%$ of its own weight) of only $\sim$50~mg. Thus, ``backpack'' control devices, tested on large free-moving insects like beetles, moths and cockroaches~\cite{Sato2008, Alper2007, Sato2009, Latif2012, Kakei2022}, are not practical for guiding the flight of small insects such as honey bees. Also open-loop control systems have been implemented by using microelectronic backpacks to stimulate invertebrates~\cite{Ma,Nguyen2020}. The lack of feedback prevents such biohybrid systems from adjusting to dynamic environments. The reliability of stimulation signals remains an issue~\cite{Webster-Wood2023}.

Our case study uses bees to improve pollination without conventional training or payloads. We employ electro-mechanical sensors and AI-powered actuators to guide bees and enhance pollination. The absence of a payload results in low sensing and actuation power consumption of $\sim$29.5-50.2~mW per bee~\cite{Ghazoul2008}. The system is also more scalable than previous works, which require surgical intervention to implant the payload. Lastly, our FI system is not restricted to a specific task (pollination) but can be extended to more applications such as explosive and chemical detection, soil quality improvement, and water quality detection.

\section{System Architecture}

%designed with Fusion Intelligence 
At the core of an FI-based IoT system lies a structured architecture for integrating NI and AI. Such integration is essential to ensure reliable communication, efficient data processing, and effective decision-making within the FI system. In this section, we discuss both the overall system architecture and the components of the NI and AI subsystems. 

\subsection{NI Subsystem}
%NI subsystem is a crucial aspect of the FI system. 
The primary function of NI (Natural Intelligence) is to interface with the physical world by collecting data and interacting with the environment. The NI subsystem includes biological entities and a suite of electromechanical sensors and actuators. Multiple entities, such as a colony of ants or a pack of dogs, can be deployed simultaneously. Each group has an array of sensors observing their responses. Biological stimulus-response pairs serve as inputs and outputs for the sensors and actuators, enabling the training of entities for various applications. Fig.~\ref{fig:fiarchflow}(a) illustrates the NI subsystem architecture.

\subsubsection{Biological Entities}
This block hosts living creatures (animals, plants, fungi, or bacteria) operating in two modes. In \textbf{Sensing}, creatures are exposed to a physical phenomenon, and their responses are recorded and transmitted to the AI subsystem. This data trains the monitoring model, enabling AI to detect changes. In \textbf{Actuation}, creatures execute actions or tasks in an environment based on AI-generated instructions. Creatures can operate in both modes simultaneously, acting as autonomous actuators under AI supervision. For example, a creature in Sensing mode can autonomously navigate to find a target source, with the AI subsystem governing overall interactions.

\subsubsection{Sensors and Actuators}
Electromechanical sensors capture the responses of living creatures to stimuli, selected based on response characteristics. For instance, infrared cameras and microphones record motion and sound. Electromechanical actuators manage their activity per user requirements, implementing AI-generated or user-specified instructions. Training mechanisms, such as food rewards, can be automated using AI to motivate and train the creatures.

\subsubsection{Control Unit}
The control unit interfaces NI with AI using a microcontroller to convert sensor outputs to digital data and AI instructions to actuation signals, as summarized in Fig.~\ref{fig:fiarchflow}(a). Additional edge devices can interface with the microcontroller via communication ports. The control unit acquires raw sensor data through standard digital protocols (I$^2$C, SPI etc.) or an ADC. An onboard DSP filters and packages this data for network transmission. The network interface also receives and decodes AI instructions, with a DAC driving actuators to induce NI to perform tasks.

\subsubsection{Power Management Unit (PMU)}
The PMU supplies energy to the NI subsystem for field deployment. It handles battery monitoring, management, and automated food dispersal to the living creatures. The control unit’s microcontroller translates AI food delivery instructions, which the PMU then executes using dedicated actuators.

\subsection{AI Subsystem}
The AI subsystem processes data from the NI subsystem, including activity patterns of living creatures and sensor data from edge devices. This data trains AI models, utilizing cloud-based computing for resource-intensive tasks. Separate models are used for monitoring and controlling NI, as shown in Fig.~\ref{fig:fiarchflow}(b). These models interpret, control, and adapt NI in real-time, creating a flexible, self-reconfigurable FI system. The AI subsystem comprises three blocks, detailed below.

\subsubsection{Data Processing}
The Data Collection block stores incoming data from the NI subsystem, while the Data Transformation block processes this data through cleaning, scaling, and normalization. Cleaning addresses missing, inconsistent, or erroneous values, using imputation techniques for missing data and correcting or removing inconsistencies. Normalizing and scaling ensure feature comparability. The Feature Extraction block selects or creates relevant features to improve AI model performance. Data is then split into training and test sets, with the training set to train the model and the test set assessing its generalization to new data. 

\subsubsection{Supervisor}
The Model Supervisor oversees AI models, enabling automated iterative training and model deployment for an autonomous FI system. It includes a Processing Unit and a Discriminator, as shown in Fig.~\ref{fig:fiarchflow}(b). Additional memory/computing resources can be incorporated into the Model Supervisor to enhance functionality. The Processing Unit generates hyperparameters and datasets using extracted features for model training and then validating them with a test dataset. Outputs from trained models are compared by the Discriminator with user configuration and observed/expected NI activity, providing feedback to optimize training processes iteratively. The Processing Unit can also utilize output from the Monitoring AI model to incorporate domain knowledge for training the Controlling AI Model.

\subsubsection{AI Models}
Evaluation metrics such as accuracy, precision, recall, or mean squared error (MSE) are used to evaluate model performance on unseen data. The resulting AI subsystem can be hosted either in the field or in the cloud depending on the specific IoT architecture. Users can configure the AI subsystem to host several AI models to train and deploy in parallel. However, the FI system requires at least one of the following models to either monitor or control the living creatures in the NI subsystem:  
\begin{itemize}
    \item \underline{Monitoring AI model}: Identifies creature actions and correlates them with physical parameters, providing insight into NI behavior. 
    \item \underline{Controlling AI model}: Generates instructions based on monitoring model output and NI data to control creature actions.
\end{itemize}
This model combination enables autonomous learning and functioning of the AI subsystem while accommodating user inputs for desired actions.

\section{Implementation}

Designing an IoT system that collects data from NI, trains the AI models, and deploys them to manage/ control NI requires careful consideration of various components and their integration. In this section we provide a process for users to use as a reference when designing an FI system according to their application requirements.

\begin{itemize}
    \item \underline{Define problem and identify physical phenomena.}~
    Begin by clearly defining the purpose and functionality of the FI system. Identify the relevant physical phenomena associated with the sensors and actuators. For example, this could involve monitoring temperature, gas concentration, humidity, or performing navigation tasks. This step guides the selection of suitable living creatures for the NI subsystem.

    \item \underline{Select living creatures.}~
    Select creatures based on their responses to specific physical phenomena relevant to the application. Evaluate stimulus-response relationships to determine suitability. A few example stimulus-response pairs are listed as follows. Stone flies perform pushups or jump in response to decreased oxygen levels, indicating water pollution. Bumblebees exhibit unique responses to scent or taste stimuli while detecting temperature change, choosing feeders based on temperature and sucrose concentration. Ants gather in groups before, during, and after earthquakes.

    \item \underline{Select sensors and actuators.}
    Integration of NI with AI requires the behavior of living creatures to be digitized before further processing. Choose sensors and actuators based on creature responses to selected physical phenomena, establishing the AI-NI feedback loop. For example, cameras monitor the movements of ants and bees, while artificial food sources such as sugar water can train them. The next step is to design the control unit for interfacing with the selected sensors and actuators. The control unit should have the necessary processing and memory elements to receive, process, and transmit signals from the sensors/actuators. It should also have networking capabilities for communicating with the AI subsystem. 

    \item \underline{Data processing and AI model training.}~
    Raw data from the control unit is analysed and processed to create datasets for the AI models. Data processing involves steps such as data cleaning, normalization, and feature engineering, as discussed earlier. The choice of machine learning algorithms for training the AI depends on the application. The monitoring models are used to understand and interpret NI actions, while the controlling models are used to generate instructions for governing NI actions based on user requests and application requirements.  

   \item \underline{AI model deployment and control.}~
   The trained and validated AI models can now interact with the NI subsystem. The AI subsystem receives real-time NI data, which is analyzed by the monitoring AI models to make intelligent real-time decisions. The controlling AI models then generate actuator control signals to induce the NI to perform the desired actions. 
   \item \underline{Monitoring and optimization.}~
    The closed-loop nature of an FI system enables it to respond to changing conditions by adapting its functionality based on AI-driven insights. Mechanisms can be implemented to monitor the system and incorporate user inputs to optimize performance. Regular updates to the AI models can further enhance the efficiency and accuracy of the FI system over time.
\end{itemize}

\begin{figure*}[ht!]
\centering
\includegraphics[width=0.85\textwidth]{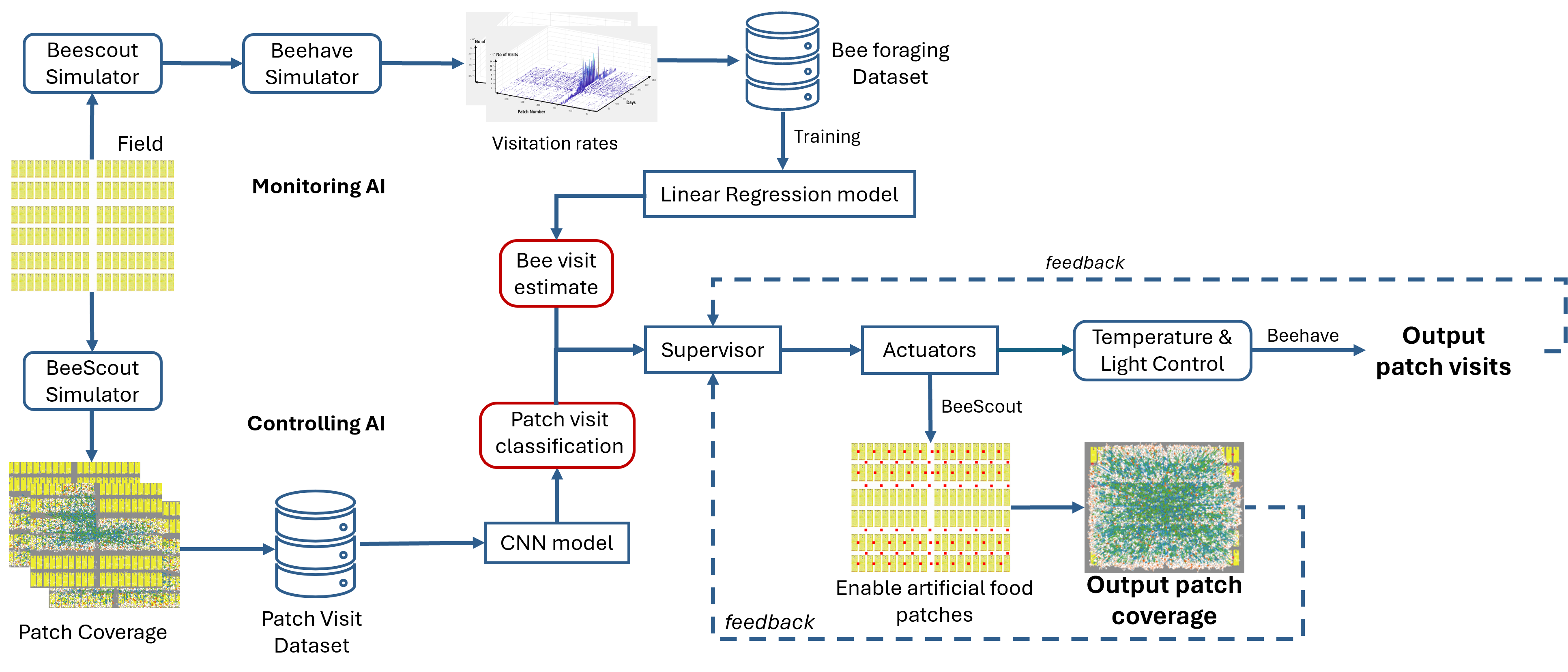}
\caption {Block diagram of the AI implemented for the FI pollination system. The monitoring model is trained to recognise patch locations that are not visited by the bees. The controlling model takes this information and adds artificial patches to guide the bees towards un-visited patches.}
\label{fig:airesult}
\end{figure*}

\section{Case study}
Ensuring adequate pollination for crops like apples, sunflowers, strawberries, and almonds is one of the challenges in farming~\cite{polli}. Insects typically perform pollination in large farms, but greenhouse and vertical farms, offering benefits like year-round production and pest protection, isolate crops from natural pollinators. Farmers augment pollination either manually or with electromechanical devices like shakers coupled with blowers. Advanced technologies like self-driving robots~\cite{bramble,robarm,YANG2023108274} and autonomous drones~\cite{dronebee,inbook} have been proposed for improving pollination. However, automation has proven to be challenging due to the complexity of tasks such as visual identification of flowers, flower manipulation, motion control, route planning, localization, and mapping~\cite{flinter,swarm}. These devices are energy-intensive. Natural pollinators are much more energy-efficient, having evolved to perform these tasks effectively.

Bumblebees carry heavier pollen loads and cover larger distances than honeybees~\cite{willmer1994superiority}. They are more resilient to cold weather, low light, forage longer, and navigate better~\cite{kevan2009measuring}. In this case study, we propose an FI system to improve and augment natural pollination using Bumblebees as sensors and actuators in the NI subsystem. Supported by the AI subsystem, FI maximizes pollination potential. We simulate this using a single virtual beehive in the Beehave suite's virtual bee foraging environment~\cite{beehave}. Beehave-BeeMapp, Beescout and Beehave-weather models were employed to simulate the foraging behaviour of in-hive bees across a given landscape~\cite{2016beescout}.

\subsection{Pollination and Foraging Activity - NI Subsystem}
Forager bees bring nectar and pollen to the hive, where it is consumed or stored in reserves. Pollination rates depend on bee foraging activity, influenced by several environmental factors from bee traits to genetics~\cite{hall2020bee}. Increased foraging leads to higher visitation metrics like "Daily Visits," "Foraging Period," "Trips per hour of Sunshine," and "Completed Foraging Trips" present in Beehave. Multiple visits to a flower can enhance pollination success ($\sim$40\% for a single visit), but subsequent visits to a pollinated flower do not improve efficacy. Rather, it is beneficial if bees could cover as many flower patches as possible, measured by "Detected Patches" and "Covered Area in km\textsuperscript{2}" in the simulator.

Quantifying changes in pollination effectiveness due to foraging is complex due to the interplay between various factors. Here we use a Pollination Improvement Index (PII) that combines field coverage and visitation metrics. Mathematically, PII is defined as
\begin{equation}
     PII = W_1({{\Delta_{PD}}})+W_2({{\Delta_{DV}}}),    
\end{equation}
where $\Delta_{PD}$ and $\Delta_{DV}$ quantify the improvements in patch detection and daily visits, respectively, while $W_1$ and $W_2$ represent the normalized weights of each factor. 
 
\subsection{Foraging Activity Monitoring and Control - AI Subsystem}
Beescout generates a foodflow file for Beehave simulation, detailing patch size, coordinates, distance from hive, nectar/pollen quantity, and detection probability. The FI system adjusts temperature and light hours in Beehave simulations using the Weather Module (2016). A modified weather file incorporates changes in foraging hours due to minimum foraging temperature (15$^{\circ}$C) and additional light hours. Beehive simulation produces daily bee visit counts per patch, forming the training dataset. A linear regression model, trained on Beehave input parameters, quantifies foraging data dependence. This monitoring AI model optimizes temperature and light hours (Fig.~\ref{fig:airesult}) to maximize daily visits, enabling the supervisor to enhance foraging activity.

The Beescout model simulated bee scouting behavior in a landscape~\cite{2016beescout}, producing a dataset of images with varying bee coverage levels over the field. A CNN as shown in Fig.~\ref{fig:airesult}, serves as the controlling AI model, classifying regions as "low," "normal," or "high" coverage. The supervisor then tags low/normal coverage regions, comparing estimated and required coverage to generate a loss function. Artificial food patches acting as actuators, guide bees to low-coverage areas. The feedback loop shown in Fig.~\ref{fig:airesult} enables the supervisor to optimize placement of food patches across the field. 

\section{Simulation Setup and Results}

\begin{figure}[t]
\centering
\includegraphics[width=1\linewidth]{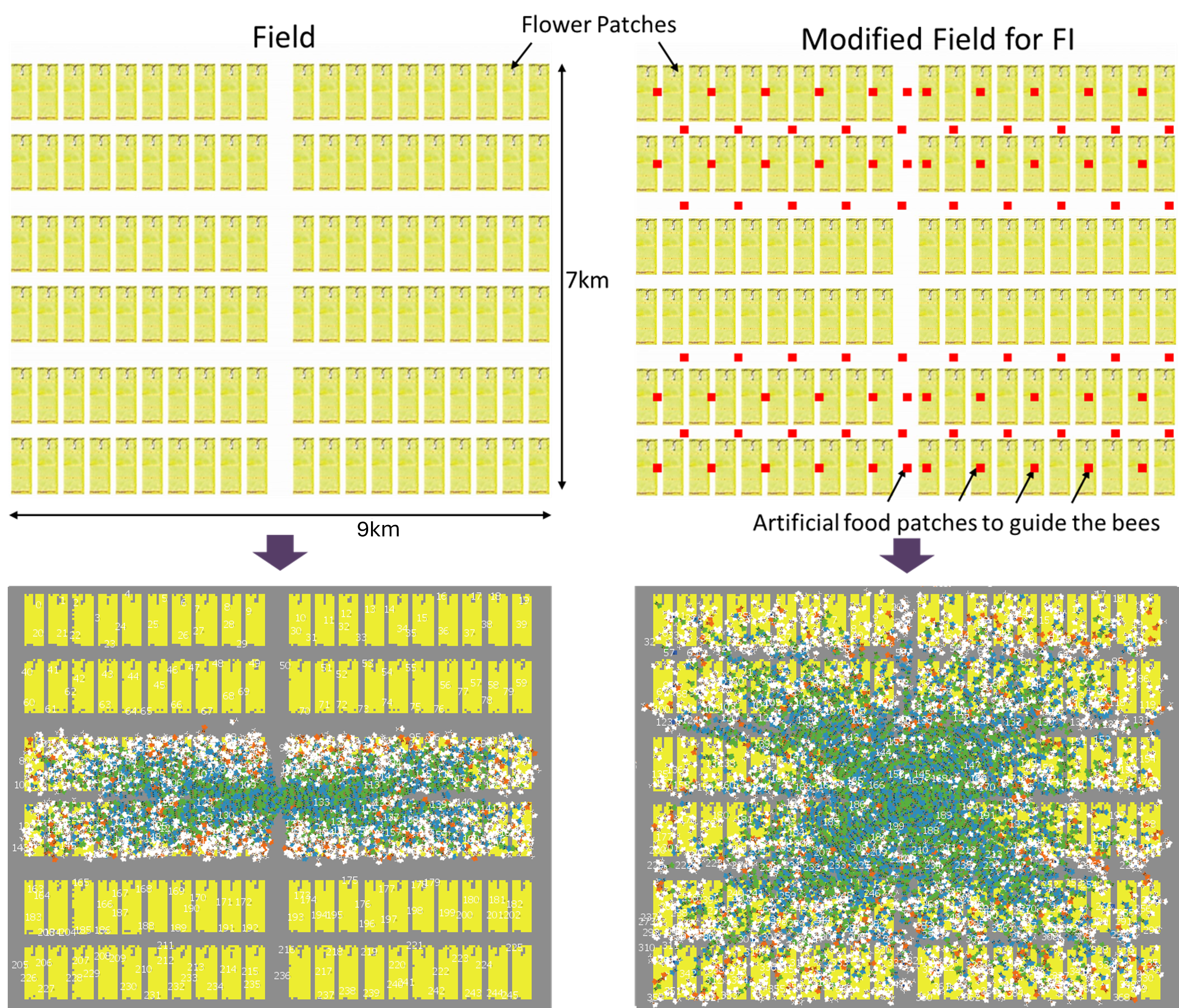}
\caption {Simulated bee foraging activity over the landscape in the baseline (left) and FI-enabled (right) scenarios. The yellow rectangular patches represent the flower/crop under cultivation. The smaller red patches represent artificial food patches placed between the crops enabled by the FI system to guide the bees.}
\label{fig:field}
\end{figure} 

\subsection{Simulation Setup}
The Beehave model, updated in 2016 with the BeeMapp module, was used for simulations along with Beehave Weather for BeeMapp (2016) and Beescout (2016) models in a NetLogo environment. The input field image for Beescout and Beehave represents a real landscape of 7,200 hectares (approx. 9~km $\times 8$~km) of farmland with a single crop (yellow) and grey gaps as physical boundaries. Beescout simulations feature a colony of 10,000 bees, typical in commercial rearing, with foraging period set to run for 9 hours, and all other options set to their default values. Beehave simulations run from January 1st to December 31st with weather conditions consistent to the location at Hertfordshire, England, during the year 2009. The simulation begins with 10,000 worker bees classified as foragers until new in-hive bees emerge. Taking into account seasonal population changes and the fact that the simulation starts without prior bee hive activity, we compare the trends in the FI-enabled and baseline scenarios from spring (April) until the end of summer (August).

\begin{figure*}[ht!]
\centering
\includegraphics[width=0.85\linewidth]{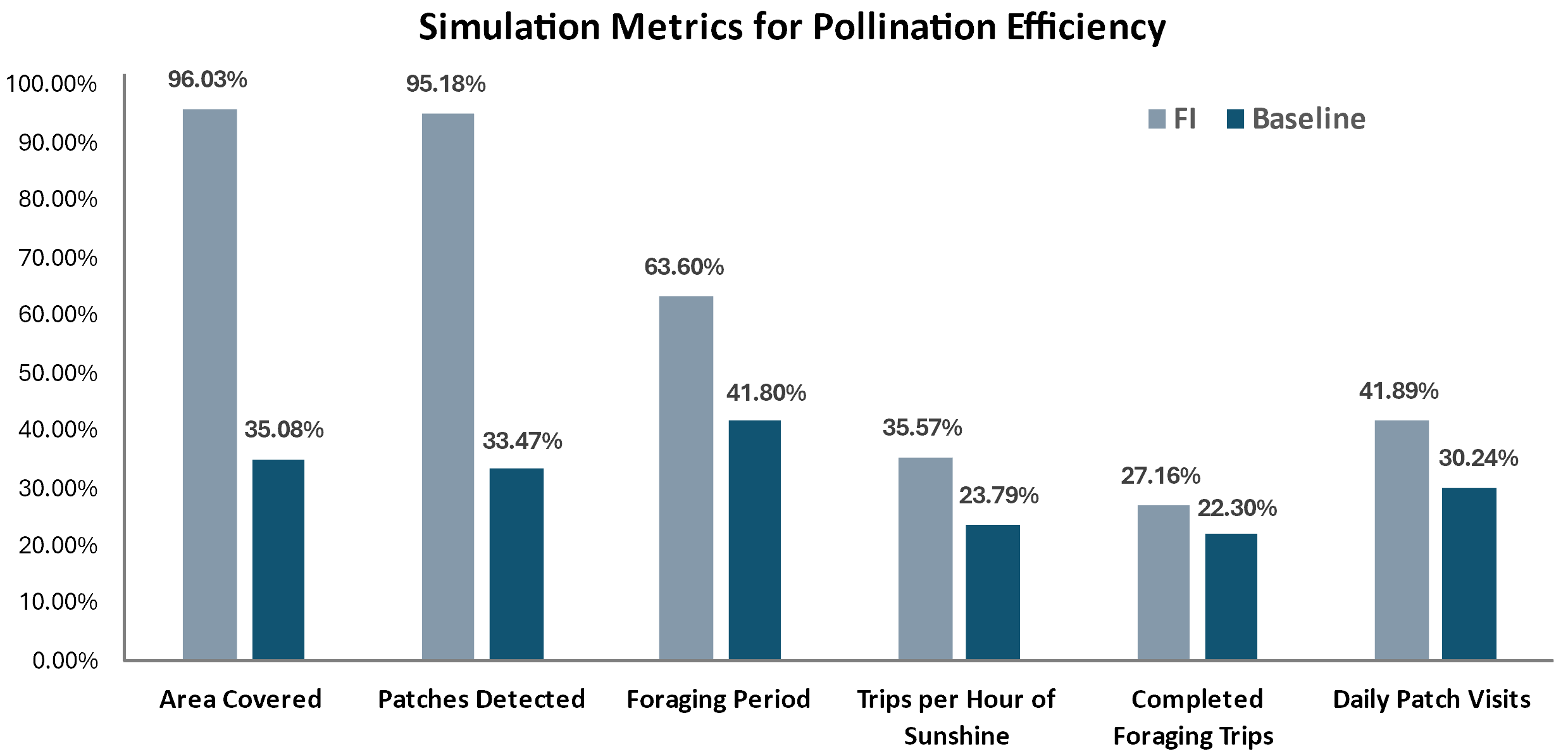}
\caption {Comparison of simulation metrics in baseline and FI-enabled scenarios for evaluating pollination efficiency. FI increased coverage of the total field area to 96\%. Similarly, the fraction of detected patches increased to 95\%. We see FI improving foraging period to 63\% taking control of the environment. The resulting improvement in visitation metrics results in average number of trips per hour and average number of completed foraging trips increasing to 35\% and 27\% respectively. Due to improved visitation, the number of total daily visits increases to 41\%.}
\label{fig:foraging}
\end{figure*} 

\subsection{Effects of FI on Pollination Efficiency}

In the baseline scenario, Beescout model coverage is limited to patches near the hive due to obstacles (grey spaces), seen in Fig.\ref{fig:field} (left panels). The FI system, with artificial food patches acting as guides, overcomes these limitations, as shown in Fig.\ref{fig:field} (right panels). Fig.~\ref{fig:foraging} compares patch coverage and visitation metrics for baseline and FI scenarios. Baseline Beescout generated 245 patches, covering 35.1\% area and discovering 33.5\% of all the patches. FI increased patches to 337 with the addition of artificial patches but these patches were removed while evaluating field coverage, thus being consistent with baseline. Area coverage and patch discovery rose to 96.0\% and 95.2\%, respectively. The monitoring regression model scored an $R^{2}$ of 88\%. The controlling CNN classification model accuracy was at 90\%. The FI system extended foraging periods to 63\% from April to August due to cold, cloudy weather in 2009. This led to total patch visits increasing to 41\% during this period.

To summarize as seen in Fig.~\ref{fig:foraging}, the FI system improved the area covered by the bees, foraging period, total number of foraging trips, and total trips per hour of available sunshine. The positive changes in these factors led to increased patch detection and daily visits, which in return improve pollination. This improvement was quantified by approximating the PII with equal weights of $W_1=W_2=0.5$, resulting in 
\begin{equation}
	PII \approx 0.5 \times ({{\Delta_{PD}}}+{{\Delta_{DV}}}).
\end{equation}
The simulation resulted values of $\Delta_{PD}=61.71$\% and $\Delta_{DV}=38$\% lead to $PII=49.85$\%. Thus, the FI system provides a pollination improvement of $\sim$50\% within the simulated landscape.

\subsection{Limitations}
The Beehave simulator lacks options to accurately represent greenhouse settings and 3D physical obstructions that could hinder bee navigation. It also only simulates a single beehive. A physical implementation of FI system would reveal challenges such as hive maintenance and ethical considerations would need to be addressed. A virtual evaluation negates the requirement to tackle these broader problems, making it suitable for initial studies of the FI paradigm. However, the lack of similar comprehensive simulation models for other insects largely limits the scope of virtual experiments to bees.

\section{Conclusion}
\label{sec:conclusion}
Replicating tasks performed by living creatures with engineered systems cannot match the efficiency of biological mechanisms evolved over millennia. Previous work has enhanced engineered systems with AI, but this paper proposes integrating AI with NI to form FI systems. This integration allows AI to leverage NI's adaptive, resilient, and unique sensing capabilities, while NI benefits from AI's computational and analytical strengths. We presented a model and architecture for an FI system, detailed its design, and demonstrated its practical application. Simulation results show significant performance improvements in agricultural scenarios. Future work on FI includes: (1) exploring alternative AI-NI cooperation mechanisms, (2) experimentally demonstrating FI's efficacy, and (3) evaluating FI's potential in various applications leveraging AI and NI's complementary capabilities.

\bibliographystyle{IEEEtran}
\bibliography{main}
\end{document}